 \documentclass[tablecaption=bottom,wcp]{jmlr} 

\usepackage{booktabs}
\usepackage{multirow}
\usepackage{bm}
\usepackage[load-configurations=version-1]{siunitx} 

\theorembodyfont{\upshape}
\theoremheaderfont{\scshape}
\theorempostheader{:}
\theoremsep{\newline}

\jmlrproceedings{AABI 2020}{3rd Symposium on Advances in Approximate Bayesian Inference, 2020}

\title[On Batch Normalisation for Approximate Bayesian Inference]{On Batch Normalisation for Approximate Bayesian Inference}

\author{\Name{Jishnu Mukhoti} \Email{jishnu@robots.ox.ac.uk}\\
 \Name{Puneet K. Dokania} \Email{puneet@robots.ox.ac.uk}\\
 \Name{Philip H.S. Torr} \Email{phst@robots.ox.ac.uk}\\
 \Name{Yarin Gal} \Email{yarin@cs.ox.ac.uk}\\
 \addr University of Oxford, Oxford, United Kingdom}

\begin{document}

\maketitle

\begin{abstract}
We study batch normalisation in the context of variational inference methods in Bayesian neural networks, such as mean-field or MC Dropout. 
We show that batch-normalisation does not affect the optimum of the evidence lower bound (ELBO). Furthermore, we study the Monte Carlo Batch Normalisation (MCBN) algorithm, proposed as an approximate inference technique parallel to MC Dropout, and show that for larger batch sizes, MCBN fails to capture epistemic uncertainty. Finally, we provide insights into what is required to fix this failure, namely having to view the mini-batch size as a variational parameter in MCBN. We comment on the asymptotics of the ELBO with respect to this variational parameter, showing that as dataset size increases towards infinity, the batch-size must increase towards infinity as well for MCBN to be a valid approximate inference technique.
\end{abstract}


\section{Introduction}
\label{sec:intro}

Deep neural networks have achieved tremendous success in several fields and are now being applied to various important real-life applications like medical diagnosis \citep{esteva2017dermatologist} and autonomous driving \citep{levinson2011towards}, where it is particularly important to produce reliable uncertainty estimates in addition to predictions. The Bayesian formalism \citep{neal2012bayesian, mackay1992bayesian} provides a principled approach for modeling uncertainty in deep neural networks. However, performing exact Bayesian inference in deep neural networks is computationally intractable. In this regard, one of the well-known and accepted methods of performing approximate Bayesian inference is variational inference (VI) \citep{graves2011practical, hinton1993keeping} in which an approximating distribution is used instead of the true Bayesian posterior over model parameters.

There have been several works \citep{graves2011practical, hernandez2015probabilistic, blundell2015weight, gal2016dropout} which have tried to apply VI in deep neural networks. Among them, \citep{gal2016dropout} provided a practical solution by showing that any neural network trained using dropout layers is performing VI and can be treated as a Bayesian neural network. Their method, MC Dropout, uses dropout at test time to sample from the approximate posterior in a deep neural network.

Following on \citep{gal2016dropout}, the paper \citep{bn_unc} claimed that batch normalisation (BN) layers can also be used to cast deep neural networks as Bayesian models. Their method, Monte Carlo Batch Normalisation (MCBN) leverages the randomness of mini-batch statistics computed by BN layers due to the stochasticity of mini-batches drawn from the training set. 

The running mini-batch statistics (mean and variance) of BN layers are stochastic estimators of the true \textbf{population statistics}. This observation has direct implications on how batch normalisation can or cannot be used to perform VI in deep neural networks. In this paper, we use this observation to first show that adding BN layers in deep neural networks does not affect the probabilistic inference of existing VI methods like MC Dropout \citep{gal2016dropout} or Bayes by Backprop \citep{blundell2015weight}. Secondly, we also expose a failure case of MCBN. In particular, we see that by increasing the mini-batch size, the epistemic uncertainty \citep{kendall2017uncertainties} estimates produced by a model reduce to zero as BN layers become deterministic. We empirically see how this renders a model completely incapable of distinguishing between in-distribution and out-of-distribution samples. 
Finally, we provide insight into what is required to fix this failure, namely having to view the mini-batch size as a variational parameter in MCBN. We comment on the asymptotics of the ELBO with respect to this variational parameter, showing that as dataset size increases towards infinity, the batch-size must increase towards infinity as well for MCBN to be a valid approximate inference technique.

\section{Batch Normalisation with Variational Inference}
\label{sec:bn_vi}

\paragraph{Background on Variational Inference (VI):}
In a Bayesian framework, we are interested in finding the posterior distribution $p(\theta|\bm{\mathrm{X, Y}})$ over model parameters $\theta$ given data $\mathcal{D} = \{\bm{\mathrm{X, Y}}\} = \{(\bm{\mathrm{x}}_i, \bm{\mathrm{y}}_i)_{i=1}^{N}\}$. However, finding the true posterior for modern neural networks is computationally intractable. Hence, in variational inference, we find an approximating parameterised distribution $q_{\phi}(\theta)$ (also called the \textit{variational distribution} with parameters $\phi$) over model weights $\theta$ to approximate the true posterior. We want the variational distribution to be as close to the true posterior as possible. This is done by minimising the KL divergence $\mathrm{KL}(q_\phi(\theta) || p(\theta|\bm{\mathrm{X, Y}})$ between $q_{\phi}(\theta)$ and the true posterior $p(\theta|\bm{\mathrm{X, Y}})$. Minimising this KL divergence is equivalent to maximising the evidence lower bound (ELBO) (or minimising its negative):
\begin{equation}
\label{eq:elbo}
\begin{split}
    \mathcal{L}_{\mathrm{ELBO}}(\phi) & = - \int q_\phi(\theta) \log p(\bm{\mathrm{Y}}|\bm{\mathrm{X}}, \theta) d\theta + \mathrm{KL}(q_\phi(\theta) || p(\theta)) \\
                                      & = - \sum_{i=1}^N \int q_\phi(\theta) \log p(\bm{\mathrm{y}}_i|\bm{\mathrm{x}}_i, \theta)d\theta + \mathrm{KL}(q_\phi(\theta) || p(\theta))
\end{split}
\end{equation}
In variational inference, we minimise $\mathcal{L}_{\mathrm{ELBO}}(\phi)$ to find the optimal variational distribution $q_{\phi}^{*}(\theta)$ which explains our data. Note that the ELBO objective ($\mathcal{L}_{\mathrm{ELBO}}(\phi)$) is over variational parameters $\phi$ and not model weights $\theta$.

\paragraph{Does Batch Normalisation (BN) impact VI?} The batch normalisation \citep{bn, lecun2012efficient} operation is used to standardise each layer's input to have zero mean and unit variance. Given input $\bm{\mathrm{h}}^l$ for a fully connected layer $l$, the BN operation uses two data-dependent statistics: the mini-batch mean $\bm{\mu}_B^l$ and the mini-batch variance $(\bm{\sigma}^2)_B^l$ to compute:
\begin{equation}
\label{eq:bn}
    \hat{\bm{\mathrm{h}}}^l = \left(\frac{\bm{\mathrm{h}}^l - \bm{\mu}_B^l}{\sqrt{(\bm{\sigma}^2)_B^l}}\right)\bm{\gamma}^l + \bm{\beta}^l
\end{equation}
where $\bm{\gamma}^l$ and $\bm{\beta}^l$ are the scale and shift parameters which are learnt through backpropagation. If we compute $\mathcal{L}_{\mathrm{ELBO}}(\phi)$ using the entire dataset $\mathcal{D} = \{(\bm{\mathrm{x}}_i, \bm{\mathrm{y}}_i)_{i=1}^{N}\}$ as shown in equation \ref{eq:elbo}, the mini-batch statistics $\{\bm{\mu}_B^l, (\bm{\sigma}^2)_B^l\}$ are exactly equal to the population statistics $\{\bm{\mu}^l, (\bm{\sigma}^2)^l\}$ and hence, are fixed. However, in practical situations, the entire dataset is not used while optimising $\mathcal{L}_{\mathrm{ELBO}}(\phi)$ and mini-batch optimisation methods \citep{hoffman2013stochastic} are employed. In such cases, the $\mathcal{\hat{L}}_{\mathrm{ELBO}}(\phi)$ computed from a minibatch $B$ is:
\begin{equation}
    \mathcal{\hat{L}}_{\mathrm{ELBO}}(\phi) = - \frac{N}{|B|} \sum_{i \in B} \int q_\phi(\theta) \log p(\bm{\mathrm{y}}_i|\bm{\mathrm{x}}_i, \theta)d\theta + \mathrm{KL}(q_\phi(\theta) || p(\theta)).
\end{equation}
During training, BN layers keep a running average of means and variances (batch statistics) which get updated with each incoming mini-batch from the training set \citep{bn}. Following the law of large numbers, in expectation, the running batch statistics converge to the population statistics $\{\bm{\mu}^l, (\bm{\sigma}^2)^l\}$ and hence, $\mathcal{\hat{L}}_{\mathrm{ELBO}}(\phi)$ computed from a mini-batch forms an unbiased estimator of $\mathcal{L}_{\mathrm{ELBO}}(\phi)$ computed using the entire dataset, i.e., $\mathbb{E}_B[\mathcal{\hat{L}}_{\mathrm{ELBO}}(\phi)] = \mathcal{L}_{\mathrm{ELBO}}(\phi)$. There is no change to the optimum of the ELBO $\mathcal{L}_\mathrm{ELBO}(\phi)$ due to the presence of BN layers. Thus, batch normalisation does not impact the probabilistic inference of other VI methods like Mean Field VI \citep{graves2011practical, blundell2015weight} or MC Dropout \citep{gal2016dropout}.

\section{Stochastic Batch Normalisation (MCBN) as VI}
\label{sec:mcbn}

\paragraph{What is MCBN?} In the work \citep{bn_unc}, the authors propose using batch normalisation as a way of performing VI in deep neural networks. In a nutshell, they categorise the parameters of a neural network into: 
\begin{enumerate}
    \item \textit{Learnable Parameters:} $\theta = \{(\bm{\mathrm{W}}^l, \bm{\beta}^l, \bm{\gamma}^l)_{l=1}^{L}\}$, weight matrices and $(\bm{\beta, \gamma})$ BN parameters learnable through backpropagation for a neural network with $L$ layers.
    
    \item \textit{Stochastic Parameters:} $w = \{(\bm{\mu}_B^l, (\bm{\sigma}^2)_B^l)_{l=1}^L\}$, the running average batch statistics (mean and variance) for all BN layers.
\end{enumerate}
In their method, they first train a network with BN layers using a conventional optimiser (like SGD or Adam etc.) and loss function (like cross entropy). During test time, they keep the BN layers in training mode and for each test sample $\bm{\mathrm{x}}^*$, they perform multiple stochastic forward passes using different, randomly chosen mini-batches from the training set. Each such forward pass corresponds to a particular instantiation of $w = \{(\bm{\mu}_B^l, (\bm{\sigma}^2)_B^l)_{l=1}^L\}$ and provides a single output $\bm{\hat{\mathrm{y}}}_t^*$ (for the $t^{th}$ stochastic forward pass, say). The mean $\mathbb{E}[\bm{\mathrm{y}}] \approx \frac{1}{T}\sum_{t=1}^{T} \bm{\hat{\mathrm{y}}}_t^*$ is then used as the prediction and some measure of variance of these outputs can be interpreted as an uncertainty estimate. It is important to note here that the mini-batch size used during stochastic forward passes at test time is the same as the mini-batch size used for training. This method was named \textit{Monte Carlo Batch Normalisation} (MCBN). 

\paragraph{Failure case of MCBN:} In MCBN, the randomness of model outputs at test time is solely dependent on the stochastic parameters $w = \{(\bm{\mu}_B^l, (\bm{\sigma}^2)_B^l)_{l=1}^L\}$, which in turn are dependent on the mini-batch size. As mentioned before, in case of larger mini-batch sizes, the batch statistics will converge to the population statistics, which is in fact desirable for BN layers following their original motivation. When the batch size equals the dataset size, there will be no stochasticity at all, and `full BN' is recovered (the optimal case for regularisation). However, in this case, any measure of epistemic uncertainty which depends on variation of model outputs over stochastic forward passes will reduce to zero and hence, the resulting model won't be able to capture epistemic uncertainty at all.

\begin{figure*}[!t]
	\centering
	\subfigure[]{\includegraphics[width=0.32\linewidth]{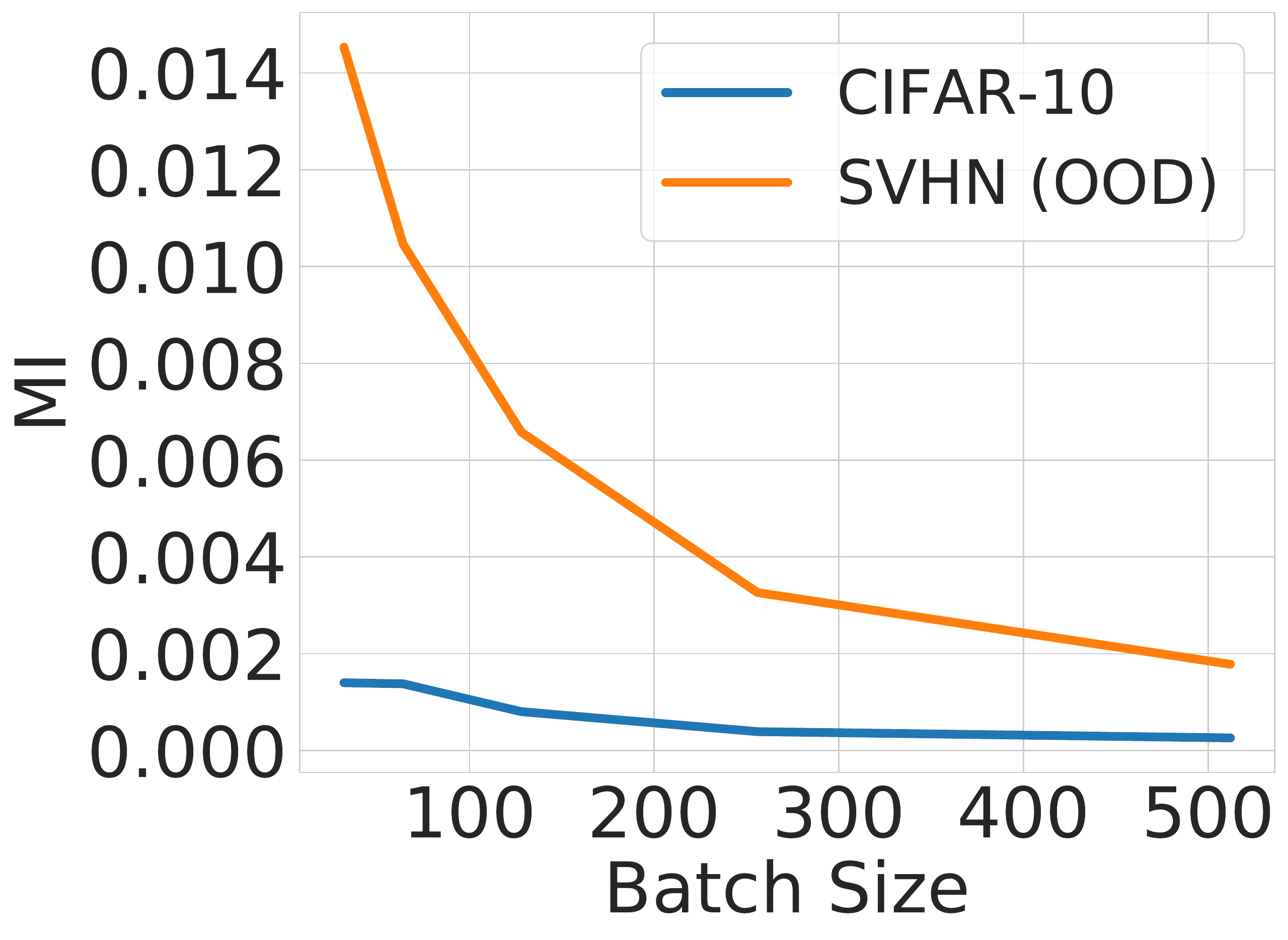}}
	\subfigure[]{\includegraphics[width=0.32\linewidth]{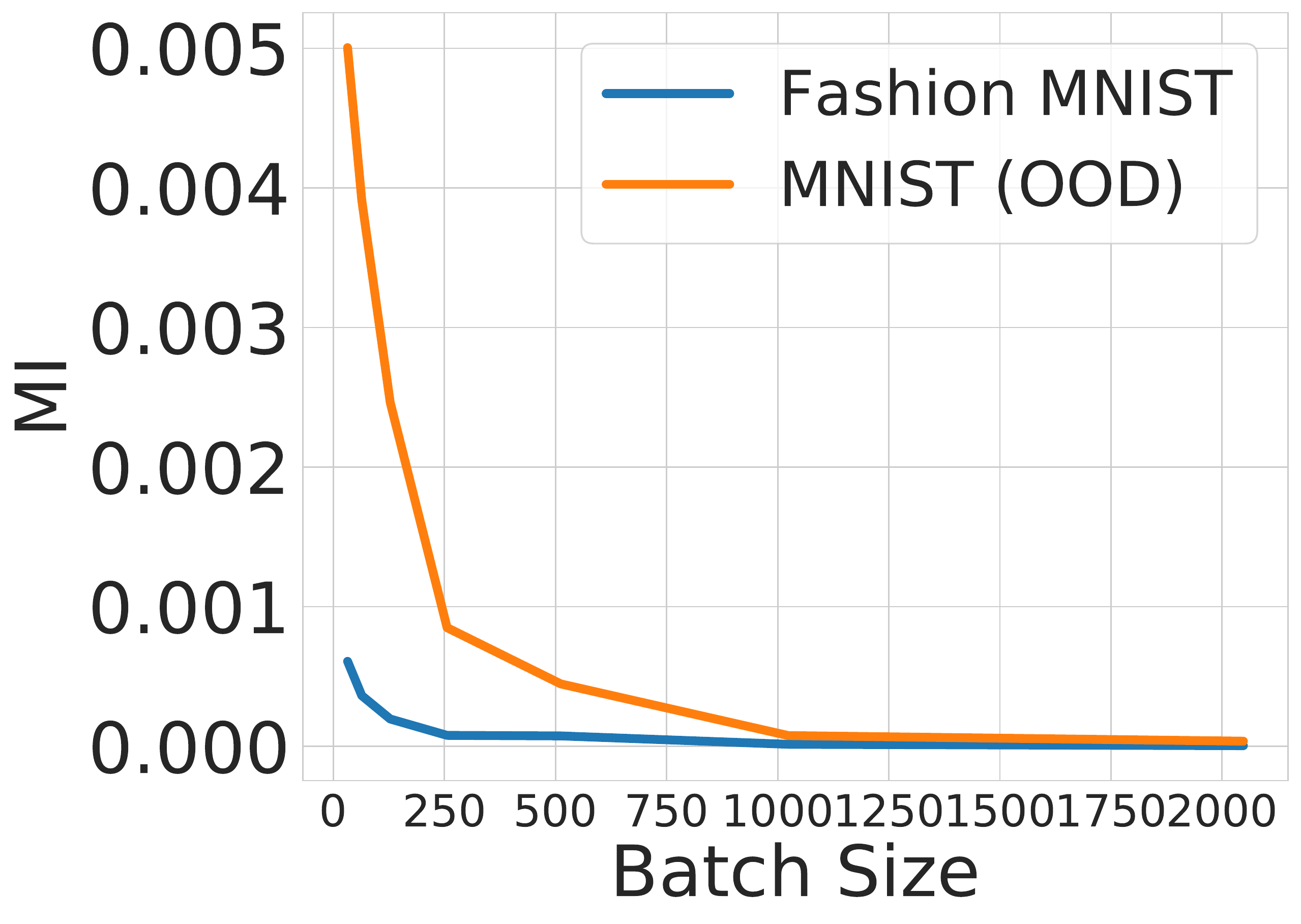}}
	\subfigure[]{\includegraphics[width=0.32\linewidth]{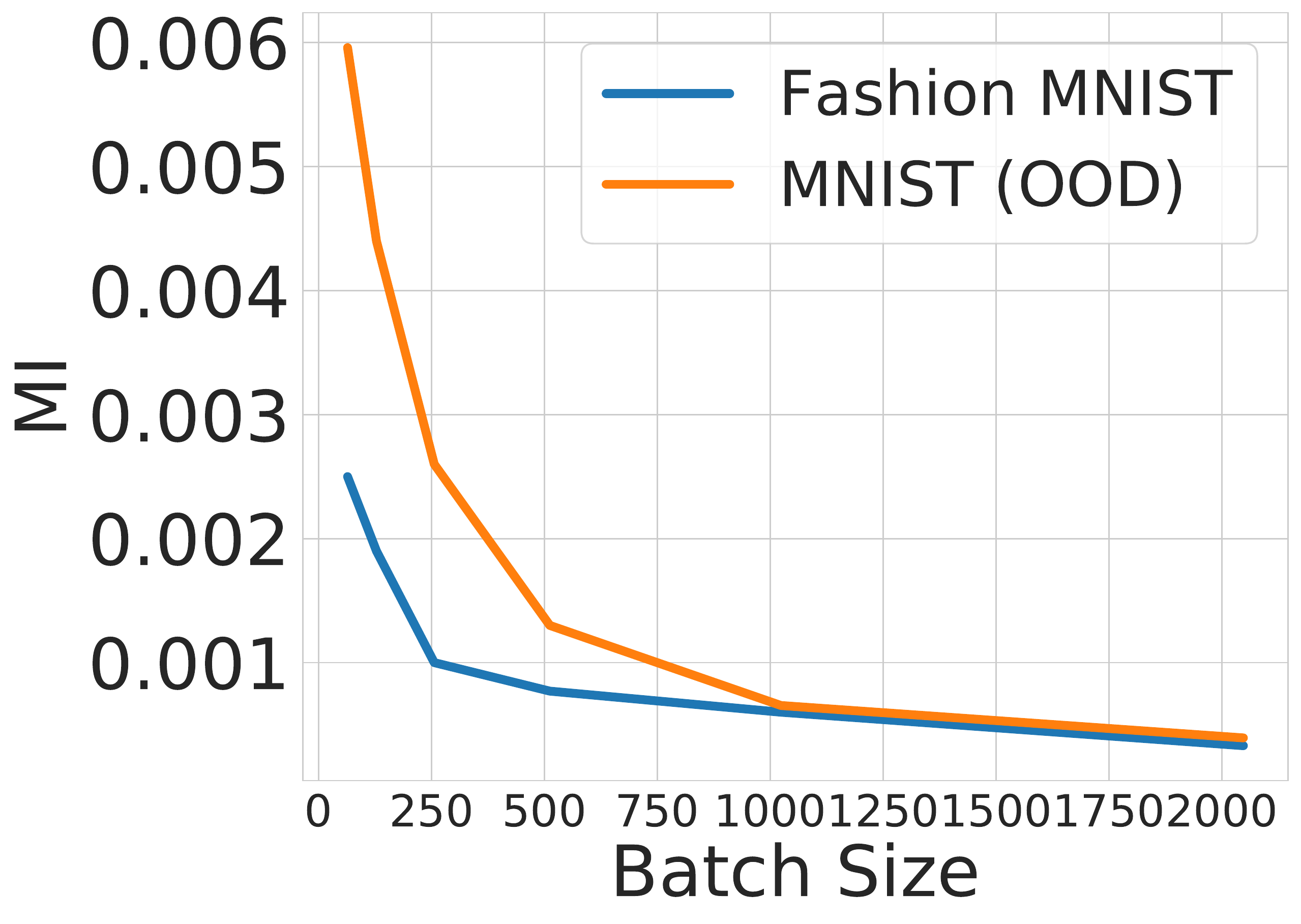}}
	\vspace{-4mm}
	\caption{Mutual Information (MI) between predictive and posterior distributions plotted against mini-batch size for (a) a ResNet-50 trained on CIFAR-10 and tested on SVHN (OOD), (b) ResNet-50 trained on Fashion MNIST and tested on MNIST (OOD) and (c) VGG-16 trained on Fashion MNIST and tested on MNIST (OOD).}
	\label{fig:bs_mi}
	\vspace{-6mm}
\end{figure*}

We empirically validate this claim by showing the effect of increasing the mini-batch size in MCBN on the epistemic uncertainty of the model. As epistemic uncertainty should be high on unseen data \citep{kendall2017uncertainties}, we choose a notably difficult pair of in-distribution and OOD datasets \citep{nalisnick2019hybrid}, (CIFAR-10 \citep{krizhevsky2009learning} vs. SVHN \citep{netzer2011reading}) and (Fashion-MNIST \citep{xiao2017fashion} vs MNIST \citep{lecun1998mnist}). We train a ResNet-50 on CIFAR-10 and Fashion-MNIST and we train a VGG-16 on Fashion-MNIST using different training batch sizes from 32 to 2048 (see Table \ref{table:error_tab} for specific batch sizes). We use SGD with a momentum of 0.9 as the optimiser and cross-entropy as the objective function. We train the ResNet-50 on CIFAR-10 for 350 epochs with an initial learning rate of 0.1 and drop the learning rate by a factor of 10 at epochs 150 and 250. For the models trained on Fashion-MNIST we use 100 training epochs and drop the learning rate at epochs 40 and 60. During test time, we use 20 stochastic forward passes to make predictions for each model.

In order to compute the epistemic uncertainty, we use the mutual information \citep{yarinthesis} between the predictive and the posterior distributions as the uncertainty estimate of choice. In Table \ref{table:error_tab}, we show the test set errors of the models for every batch size. Note that we did not train ResNet-50 on CIFAR-10 with batch sizes 1024 and 2048 due to compute constraints. Finally, we show the variation of epistemic uncertainty with batch size for both in-distribution and OOD datasets in Figure \ref{fig:bs_mi}.

It is very clear from Figure \ref{fig:bs_mi} that \textit{with increase in batch sizes, the model's epistemic uncertainty lowers to the point where it can no longer distinguish between in and out of distribution samples}. In fact, for the models trained on Fashion MNIST, we find the epistemic uncertainty to almost become 0 for batch sizes 1024 and 2048. Thus, although the test set accuracy remains more or less the same, MCBN fails to capture epistemic uncertainty with growing mini-batch size.

\begin{table*}[!t]
\renewcommand{\arraystretch}{1.3}
\centering
\footnotesize
\resizebox{\linewidth}{!}{%
\begin{tabular}{cccccccccc}
\toprule

\multirow{2}{*}{\textbf{Architecture}} & \multirow{2}{*}{\textbf{Dataset}} & \multirow{2}{*}{\textbf{OOD Dataset}} & \multicolumn{7}{c}{\textbf{Mini-batch size}} \\
& & & \textbf{32} & \textbf{64} & \textbf{128} & \textbf{256} & \textbf{512} & \textbf{1024} & \textbf{2048} \\
\midrule
\multirow{2}{*}{ResNet-50} & CIFAR-10 & SVHN & $0.9366$ & $0.9459$ & $0.9517$ & $0.9547$ & $0.9545$ & - & - \\
                           & Fashion-MNIST & MNIST & $0.9343$ & $0.9377$ & $0.9386$ & $0.9396$ & $0.9359$ & $0.9365$ & $0.9387$ \\
VGG-16 & Fashion-MNIST & MNIST & $0.9429$ & $0.9436$ & $0.9437$ & $0.9427$ & $0.9438$ & $0.9450$ & $0.9486$ \\
\bottomrule
\end{tabular}}
\caption{Test set accuracy computed for different MCBN mini-batch sizes. \vspace{-3mm}}
\label{table:error_tab}
\end{table*}

\paragraph{Insight into what is required to fix this failure:} The model's stochasticity in MCBN is clearly dependent on the mini-batch size used for training. Hence, the mini-batch size itself should be considered as a variational parameter much like the dropout rate is a variational parameter in MC Dropout \citep{gal2016dropout}. One can find the best dropout rate either by learning it directly \citep{gal2017concrete} or performing grid-search over possible dropout rates \citep{mukhoti2018importance}. Similarly, one could think of performing a grid-search over possible mini-batch sizes and find the one which maximises the 
ELBO.
\paragraph{Comment on ELBO asymptotics:} 
We briefly comment on the asymptotics of the ELBO with respect to this variational parameter. As the dataset size increases towards infinity, we expect epistemic uncertainty of a parametric model to diminish to zero. However, as we commented above, for the epistemic uncertainty of MCBN to diminish to zero, the batch-size must increase towards infinity as well. This may be impractical in certain applications.

\section{Conclusion}
In this paper we have studied how batch normalisation layers fit into the context of variational inference in deep neural networks. Mini-batch statistics in BN layers are stochastic estimators of the population statistics. We use this observation to show that batch normalisation does not affect the probabilistic inference of VI approaches. Furthermore, we use the same observation to expose a failure case of the MCBN \citep{bn_unc} algorithm and see that for bigger batch sizes, MCBN is unable to produce sensible epistemic uncertainty and cannot distinguish in and out of distribution samples. Finally, we provide some insights into how to fix the problem and leave the development of more involved approaches on MCBN as future work.


\bibliography{jmlr-sample}

\end{document}